\documentclass[11pt,a4paper]{article}
\usepackage[hyperref]{acl2017}
\usepackage{times}
\usepackage{latexsym}
\usepackage{graphicx}
\usepackage{hyperref}
\usepackage{parskip}

\hypersetup{
  colorlinks=true,
  citecolor=blue
}

\aclfinalcopy

\title{A Survey of Neural Network Techniques\\ for Feature Extraction from Text}

\author{
  Vineet John \\
  University of Waterloo \\
  {\tt v2john@uwaterloo.ca} \\
}

\date{}

\begin{document}

\maketitle

\begin{abstract}
  This paper aims to catalyze research discussions about text feature extraction techniques using neural network architectures.
  The research questions discussed here focus on the state-of-the-art neural network techniques that have proven to be useful tools for language processing, language generation, text classification and other computational linguistics tasks.
\end{abstract}

\section{Motivation} 
\label{sec:motivation}

  A majority of the methods currently in use for text-based feature extraction rely on relatively simple statistical techniques. For instance, a word co-occurrence model like n-grams or a bag-of-words model like TF-IDF.

  The motivation of this research project is to identify and survey the techniques that use neural networks and study them in juxtaposition with the traditional text feature extraction models to show their differences in approach.

  Feature extraction of text can be used for a multitude of applications including - but not limited to - unsupervised semantic similarity detection, article classification and sentiment analysis.

  The goal of this project is to document of the differences, advantages and drawbacks in the domain of feature extraction from text data using neural networks. It also sketches the evolution of such techniques over time.

  This report could serve as a quick cheat-sheet for engineers looking to build a text classification or regression pipeline, as the discussion (Section \ref{sec:discussion}) would serve to map a use-cases to feature extraction implementation specifics.


\section{Research Questions} 
\label{sec:research_questions}

  \begin{itemize}
    \item [\textbf{RQ1}]
    What are the relatively simple statistical techniques to extract features from text?
    \item [\textbf{RQ2}]
    Is there any inherent benefit to using neural networks as opposed to the simple methods?
    \item [\textbf{RQ3}]
    What are the trade-offs that neural networks incur as opposed to the simple methods?
    \item [\textbf{RQ4}]
    How do the different techniques compare to each other in terms of performance and accuracy?
    \item [\textbf{RQ5}]
    In what use-cases do the trade-offs outweigh the benefits of neural networks?
  \end{itemize}


\section{Methodology} 
\label{sec:methodology}

  The research questions listed in Section~\ref{sec:research_questions} will be tackled by surveying a few of the important overview papers on the topic\cite{goldberg2016primer}\cite{bengio2003neural}\cite{morin2005hierarchical}. A few of the groundbreaking research papers in this area will also be studied, including word embeddings\cite{mikolov2013efficient}\cite{mikolov2013distributed}\cite{mikolov2013linguistic}.

  In addition to this, other less-obvious methods of features extraction will be surveyed, including tasks like part-of-speech tagging, chunking, named entity recognition, and semantic role labeling\cite{socher2011parsing}\cite{luong2013better}\cite{maas2015lexicon}\cite{li2015hierarchical}\cite{collobert2011natural}\cite{pennington2014glove}.


\section{Background} 
\label{sec:background}

  This section provides a high level background of the tasks within Computational Linguistics.

  \subsection{Part-of-Speech Tagging} 
  \label{sub:part_of_speech_tagging}

    \begin{itemize}
      \item
      POS tagging aims to label each word with a unique tag that indicates its syntactic role, like noun, verb, adjective etc.
      \item
      The best POS taggers are based on classifiers trained on windows of text, which are then fed to a bidirectional decoding algorithm during inference.
      \item
      In general, models resemble a bi-directional dependency network, and can be trained using a variety of methods including support vector machines and bi-directional Viterbi decoders.
    \end{itemize}


  \subsection{Chunking} 
  \label{sub:chunking}

    \begin{itemize}
      \item
      Chunking aims to label segments of a sentence with syntactic constituents such as noun or verb phrases. It is also called shallow parsing and can be viewed as a generalization of part-of-speech tagging to phrases instead of words.
      \item
      Implementations of chunking usually require an underlying POS implementation, after which the words are compounded or chunked by concatenation.
    \end{itemize}


  \subsection{Named Entity Recognition} 
  \label{sub:named_entity_recognition}

    \begin{itemize}
      \item
      NER labels atomic elements in a sentence into categories such as “PERSON” or “LOCATION”.
      \item
      Features to train NER classifiers include POS tags, CHUNK tags, prefixes and suffixes, and large lexicons of the labeled entities.
    \end{itemize}


  \subsection{Semantic Role Labeling} 
  \label{sub:semantic_role_labeling}

    \begin{itemize}
      \item
      SRL aims to assign a semantic role to a syntactic constituent of a sentence.
      \item
      State-of-the-art SRL systems consist of several stages: producing a parse tree, identifying which parse tree nodes represent the arguments of a given verb, and finally classifying these nodes to compute the corresponding SRL tags.
      \item
      SRL systems usually entail numerous features like the parts of speech and syntactic labels of words and nodes in the tree, the syntactic path to the verb in the parse tree, whether a node in the parse tree is part of a noun or verb phrase etc.
    \end{itemize}



\section{Document Vectorization} 
\label{sec:document_vectorization}

  Document vectorization is needed to convert text content into a numeric vector representation that can be utilized as features, which can then be used to train a machine learning model on. This section talks about a few different statistical methods for computing this feature vector\cite{SemEvalPaper}.

  \subsection{N-gram Model} 
  \label{sub:n_gram_model}
    N-grams are contiguous sequences of `n' items from a given sequence of text or speech. Given a complete corpus of documents, each tuple of `n' grams, either characters or words are represented by a unique bit in a bit vector, which, when aggregated for a body of text, form a sparse vectorized representation of the text in the form of n-gram occurrences.


  \subsection{TF-IDF Model} 
  \label{sub:tf_idf_model}
    Term frequency - inverse document frequency (TF-IDF), is a numerical statistic that is intended to reflect how important a word is to a document in a collection or corpus \cite{sparck1972statistical}. The TF-IDF value increases proportionally to the number of times a word appears in the document, but is offset by the frequency of the word in the corpus, which helps to adjust for the fact that some words appear more frequently in general. It is a bag-of-words model, and doesn't preserve word ordering.


  \subsection{Paragraph Vector Model} 
  \label{sub:paragraph_vectors_doc2vec}

    A Paragraph Vector model is comprised of an unsupervised learning algorithm that learns fixed-size vector representations for variable-length pieces of texts such as sentences and documents \cite{le2014distributed}. The vector representations are learned to predict the surrounding words in contexts sampled from the paragraph. 

    Two distinct implementations have gained prominence in the community.
    \begin{itemize}
      \item
        Doc2Vec: A Python library implementation in Gensim. \footnote{https://radimrehurek.com/gensim/models/doc2vec.html}.
      \item
        FastText: A standalone implementation in C++. \cite{bojanowski2016enriching} \cite{joulin2016bag}.
    \end{itemize}



\section{A Primer of Neural Net Models for NLP\cite{goldberg2016primer}} 
\label{sec:a_primer_of_neural_net_models_for_nlp}

  \begin{itemize}
    \item
    Fully connected feed-forward neural networks are non-linear learners that can be used as a drop-in replacement wherever a linear learner is used.
    \item
    The high accuracy observed in experimental results is a consequence of this non-linearity along with the availability of pre-trained word embeddings.
    \item
    Multi-layer feed-forward networks can provide competitive results on sentiment classification and factoid question answering
    \item
    Convolutional and pooling architecture show promising results on many tasks, including document classification, short-text categorization, sentiment classification, relation type classification between entities, event detection, paraphrase identification, semantic role labeling, question answering, predicting box-office revenues of movies based on critic reviews, modeling text interestingness, and modeling the relation between character-sequences and part-of-speech tags.
    \item
    Convolutional and pooling architectures allow us to encode arbitrarily large items as fixed size vectors capturing their most salient features, but, they do so by sacrificing most of the structural information.
    \item
    Recurrent and recursive networks allows using sequences and trees and preserve the structural information.
    \item
    Recurrent models have been shown to produce very strong results for language modeling as well as for sequence tagging, machine translation, dependency parsing, sentiment analysis, noisy text normalization, dialog state tracking, response generation, and modeling the relation between character sequences and part-of-speech tags.
    \item
    Recursive models were shown to produce state-of-the-art or near state-of-the-art results for constituency and dependency parse re-ranking, discourse parsing, semantic relation classification, political ideology detection based on parse trees, sentiment classification, target-dependent sentiment classification and question answering.
    \item
    Convolutional nets are observed to to work well for summarization related tasks, just as recurrent/recursive nets work well for language modeling tasks.
  \end{itemize}


\section{A Neural Probabilistic Language Model} 
\label{sec:a_neural_probabilistic_language_model}

  \textbf{Goal:}
  Knowing the basic structure of a sentence, one should be able to create a new sentence by replacing parts of the old sentence with interchangeable entities\cite{bengio2003neural}.\\

  \textbf{Challenge:}
  The main bottleneck is computing the activations of the output layer, since it is a fully-connected softmax activation layer. \\

  \textbf{Description:}
  \begin{itemize}
    \item
    One of the major contributions of this paper in terms of optimizations was data parallel processing (different processors working on a different subsets of data) and asynchronous processor usage of shared memory.
    \item
    The authors propose to fight the curse of dimensionality by learning a distributed representation for words which allows each training sentence to inform the model about an exponential number of semantically neighboring sentences.
    \item
    A fundamental problem that makes language modeling and other learning problems difficult is the curse of dimensionality. It is particularly obvious in the case when one wants to model the joint distribution between many discrete random variables (such as words in a sentence, or discrete attributes in a data-mining task).
    \item
    State-of-the art results are typically obtained using trigrams.
    \item
    Language generation via substitution of semantically similar language constructs of existing sentences can be done via shared-parameter multi-layer neural networks.
    \item
    The objective of this paper is to obtain real-valued vector sequences of words and learn a joint probability function for those sequences of words alongside the feature vector, and hence, jointly learn both the real-valued vector representation and the parameters of the probability distribution.
    \item
    This probability function can be tuned in order to maximize log-likelihood of the training data, while penalizing the cost function, similar to the penalty term one used in Ridge regression.
    \item
    This will ensure that semantically similar words end up with an almost equivalent feature vectors, called learned distributed feature vectors.
    \item
    A challenge with modeling discrete variables like a sentence structure as opposed to a continuous value is that the continuous valued function can be assumed to have some form of locality, but the same assumption cannot be made in case of discrete functions.
    \item
    N-gram models try to achieve a statistical modeling of languages by calculating the conditional probabilities of each possible word that can follow a set of $n$ preceding words.
    \item
    New sequences of words can be generated by effectively gluing together the popular combinations i.e. n-grams with very high frequency counts.
  \end{itemize}


\section{Hierarchical Probabilistic Neural Network Language Model} 
\label{sec:hierarchical_probabilistic_neural_network_language_model}

  \textbf{Goal:}
  Implementing a  hierarchical decomposition of the conditional probabilities that yields a speed-up of about 200 both during training and recognition. The hierarchical decomposition is a binary hierarchical clustering constrained by the prior knowledge extracted from the WordNet\footnote{https://wordnet.princeton.edu/} semantic hierarchy\cite{morin2005hierarchical}.\\

  \textbf{Description:}
  \begin{itemize}
    \item
    Similar to the previous paper, attempts to tackle the `curse of dimensionality' (Section \ref{sec:a_neural_probabilistic_language_model}) and attempts to produce a much faster variant.
    \item
    Back-off n-grams are used to learn a real-valued vector representation of each word.
    \item
    The word embeddings learned are shared across all the participating nodes in the distributed architecture.
    \item
    A very important component of the whole model is the choice of the words binary encoding, i.e. of the hierarchical word clustering. In this paper the authors combine empirical statistics with prior knowledge from the WordNet resource.
  \end{itemize}


\section{A Hierarchical Neural Autoencoder for Paragraphs and Documents} 
\label{sec:a_hierarchical_neural_autoencoder_for_paragraphs_and_documents}

  \textbf{Goal:}
  Attempts to build a paragraph embedding from the underlying word and sentence embeddings, and then proceeds to encode the paragraph embedding in an attempt to reconstruct the original paragraph\cite{li2015hierarchical}.\\

  \textbf{Description:}
  \begin{itemize}
    \item
    The implementation uses an LSTM layer to convert words into a vector representation of a sentence. A subsequent LSTM layer converts multiple sentences into a paragraph.
    \item
    For this to happen, we need to preserve, syntactic, semantic and discourse related properties while creating the embedded representation.
    \item
    Hierarchical LSTM utilized to preserve sentence structure.
    \item
    Parameters are estimated by maximizing likelihood of outputs given inputs, similar to standard sequence-to-sequence models.
    \item
    Estimates are calculated using softmax functions to maximize the likelihood of the constituent words.
    \item
    Attention models using the hierarchical autoencoder could be utilized for dialog systems, since it explicitly models for discourse.
  \end{itemize}


\section{Linguistic Regularities in Continuous Space Word Representations} 
\label{sec:linguistic_regularities_in_continuous_space_word_representations}

  \textbf{Goal:}
  In this paper, the authors examine the vector-space word representations that are implicitly learned by the input-layer weights. These representations are surprisingly good at capturing syntactic and semantic regularities in language, and that each relationship is characterized by a relation-specific vector offset. This allows vector-oriented reasoning based on the offsets between words\cite{mikolov2013linguistic}. This is one of the seminal papers that led to the creation of Word2Vec, which is a state-of-the-art word embedding tool\cite{mikolov2013efficient}.\\

  \textbf{Description:}
  \begin{itemize}
    \item
    A defining feature of neural network language models is their representation of words as high dimensional real-valued vectors.
    \item
    In this model, words are converted via a learned lookup-table into real valued vectors which are used as the inputs to a neural network.
    \item
    One of the main advantages of these models is that the distributed representation achieves a level of generalization that is not possible with classical n-gram language models.
    \item
    The word representations in this paper are learned by a recurrent neural network language model.
    \item
    The input vector $w(t)$ represents input word at time $t$ encoded using 1-of-N coding, and the output layer $y(t)$ produces a probability distribution over words. The hidden layer $s(t)$ maintains a representation of the sentence history. The input vector $w(t)$ and the output vector $y(t)$ have dimensionality of the vocabulary.
    \item
    The values in the hidden and output layers are computed as follows:
    $$s(t) = f(Uw(t) + Ws(t-1))$$
    $$y(t) = g(Vs(t))$$
    where
    $f(z) = \frac{1}{1 + e^{-z}}$ and $g(z_m) = \frac{e^{z_m}}{\sum_k e^{z_k}} $
    \begin{figure}[ht]
      \centering
      \includegraphics[width=.4\textwidth]{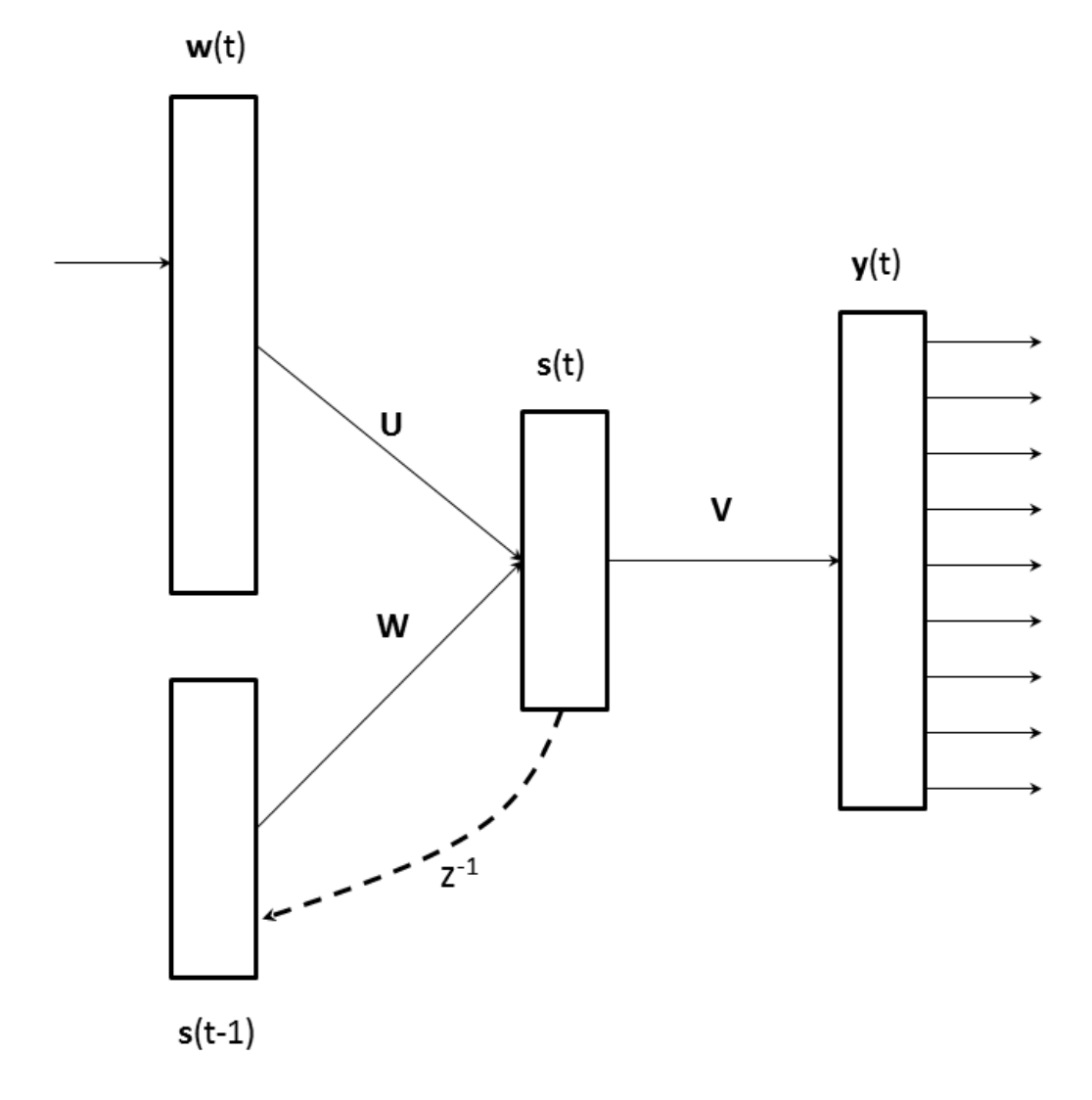}
      \caption{RNN Language Model}
      \label{fig:rnn-lang-model}
    \end{figure}
    \item
    One of the biggest features of having real-valued feature representations is the ability to compute the answer to an analogy question $a:b; c:d$ where $d$ is unknown. With continuous space word representations, this becomes as simple as calculating
    $$y = x_b - x_a + x_c$$
    $y$ is the best estimate of $d$ that the model could compute. If there is no vector amongst the trained words such that $y == x_w$, the nearest vector representation can be estimated using cosine similarity.
    $$w^* = argmax_w \frac{x_w y}{||x_w|| ||y||}$$
  \end{itemize}


\section{Better Word Representations with Recursive Neural Networks for Morphology} 
\label{sec:better_word_representations_with_recursive_neural_networks_for_morphology}

  \textbf{Goal:}
  The paper aims to address the inaccuracy in vector representations of complex and rare words, supposedly caused by the lack of relation between morphologically related words\cite{luong2013better}.\\

  \textbf{Description:}
  \begin{itemize}
    \item
    The authors treat each morpheme as a basic unit in the RNNs and construct representations for morphologically complex words on the fly from their morphemes. By training a neural language model (NLM) and integrating RNN structures for complex words, they utilize contextual information to learn morphemic semantics and their compositional properties.
    \item
    Discusses a problem that the Word2Vec syntactic relations like $$x_{apples} - x_{apple} \approx x_{cars} - x_{car}$$ might not hold true if the vector representation of a rare word is inaccurate to begin with.
    \item
    \texttt{morphoRNN} operates at the morpheme level rather than the word level. An example of the this is illustrated in Figure \ref{fig:rnn-morphology}.
    \begin{figure}[ht]
      \centering
      \includegraphics[width=.4\textwidth]{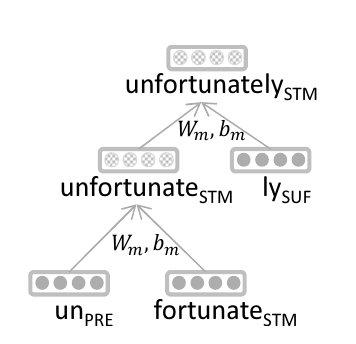}
      \caption{morphoRNN}
      \label{fig:rnn-morphology}
    \end{figure}
    \item
    Parent words are created by combining a stem vector and an affix vector, as shown in Equation \ref{eqn:parent-vector}.
    \begin{equation} \label{eqn:parent-vector}
      p = f (W_m [x_{stem} ; x_{affix}] + b_m)
    \end{equation}
    \item
    The cost function is expression in terms of the squared Euclidean loss between the newly constructed representation $p_c(x_i)$ and the reference representation $p_r(x_i)$. The cost function is given in Equation \ref{eqn:cost-function-morphornn}.
    \begin{equation} \label{eqn:cost-function-morphornn}
      J(\theta) = \sum_{i=1}^N (|| p_r(x_i) - p_c(x_i) ||^2_2) + \frac{\lambda}{2} ||\theta||^2_2
    \end{equation}
    \item
    The paper describes both context sensitive and insensitive versions of the Morphological RNN.
    \item
    Similar to a typical RNN, the network is trained by computing the activation functions and propagating the errors backward in a forward-backward pass architecture.
    \item 
    This RNN model performs better than most of the other neural language models, and could be used to supplement word vectors.
  \end{itemize}


\section{Efficient Estimation of Word Representations in Vector Space} 
\label{sec:efficient_estimation_of_word_representations_in_vector_space}

  \textbf{Goal:}
  The main goal of this paper is to introduce techniques that can be used for learning high-quality word vectors from huge data sets with billions of words, and with millions of words in the vocabulary\cite{mikolov2013efficient}.\\

  \textbf{Challenge:}
  The complexity that arises at the fully-connected output layer of the neural network is the dominant part of the computation. A couple of methods suggested to mitigate this is to use hierarchical versions of the softmax output activation units, or to refrain from performing normalization at the final layer altogether.\\

  \textbf{Description:}
  \begin{itemize}
    \item 
    The ideas presented in this paper build on the previous ideas presented by \cite{bengio2003neural}.
    \item 
    The objective was to obtain high-quality word embeddings that capture the syntactic and semantic characteristics of words in a manner that allows algebraic operations to proxy the distances in vector space.
    $$man - woman = king - queen$$ or $$tell - told = walk - walked$$
    \item
    The training time here scales with the dimensionality of the learned feature vectors and not on the volume of training data.
    \item
    The approach attempts to find a distributed vector representation of values as opposed to a continuous representation of values as computed by methods like LSA and LDA.
    \item
    The models are trained using stochastic gradient descent and backpropagation.
    \item
    The RNN models are touted to have an inherently better representation of sentence structure for complex patterns, without the need to specify context length.
    \item
    To allow for the distributed training of the data, the framework DistBelief was used with multiple replicas of the model. Adagrad was utilized for asynchronous gradient descent.
    \item
    Two distinct models were conceptualized for the training of the word vectors based on context, both of which are continuous and distributed representations of words. These are illustrated in Figure \ref{fig:cbow-skipgram}.
    \begin{figure}[ht]
      \centering
      \includegraphics[width=.4\textwidth]{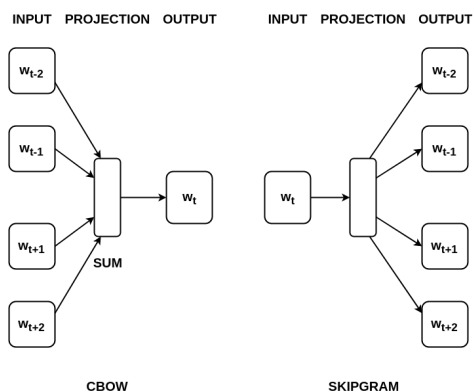}
      \caption{CBOW and Skip-gram models}
      \label{fig:cbow-skipgram}
    \end{figure}
    \begin{itemize}
      \item
      Continuous Bag-of-Words model: This model uses the context of a word i.e. the words that precede and follow it, to predict the current word.
      \item
      Skip-gram model: This model uses the current word to predict the context it appeared in.
    \end{itemize}
  \end{itemize}

  The experimental results show that the CBOW and skip-gram models consistently out-perform the then state-of-the-art models. It was also observed that after a point, increasing the dimensions and the size of the data began providing diminishing returns.


\section{Distributed Representations of Words and Phrases and their Compositionality} 
\label{sec:distributed_representations_of_words_and_phrases_and_their_compositionality}

  \textbf{Goal:}
  This paper builds upon the idea of the Word2Vec skip-gram model, and presents optimizations in terms of quality of the word embeddings as well as speed-ups while training. It also proposes an alternative to the hierarchical softmax final layer, called negative sampling\cite{mikolov2013distributed}.\\

  \textbf{Description:}
  \begin{itemize}
    \item
    One of the optimizations suggested is to sub-sample the training set words to achieve a speed-up in model training.
    \item
    Given a sequence of training words $[w_1 , w_2 , w_3 , ... , w_T]$, the objective of the skip-gram model is to maximize the average log probability shown in Equation \ref{eqn:skipgram-log-probability}
    \begin{equation} \label{eqn:skipgram-log-probability}
      \frac{1}{T} \sum_{t=1}^T \sum_{-c \leq j \leq c; j \neq 0} \log P(w_{t+j}, w_t)
    \end{equation}
    where $c$ is the window or context surrounding the current word being trained on. 
    \item 
    As introduced by \cite{morin2005hierarchical}, a computationally efficient approximation of the full softmax is the hierarchical softmax. The hierarchical softmax uses a binary tree representation of the output layer with the W words as its leaves and, for each node, explicitly represents the relative probabilities of its child nodes. These define a random walk that assigns probabilities to words.
    \item 
    The authors use a binary Huffman tree, as it assigns short codes to the frequent words which results in fast training. It has been observed before that grouping words together by their frequency works well as a very simple speedup technique for the neural network based language models.
    \item
    Noise Contrastive Estimation (NCE), which is an alternative to hierarchical softmax,  posits that a good model should be able to differentiate data from noise by means of logistic regression.
    \item
    To counter the imbalance between the rare and frequent words, we used a simple sub-sampling approach: each word within the training set is discarded with probability computed by the below formula.
    $$P(w_i) = 1 - \sqrt{\frac{t}{f(w_i)}} $$
    This is similar to a dropout of neurons from the network, except that it is statistically more likely that frequent words are removed from the corpus by virtue of this method.
    \item
    Discarding the frequently occurring words allows for a reduction in computational and memory cost.
    \item
    The individual words can easily be coalesced into phrases using unigram and bigram frequency counts, as shown below.
    $$score(w_i, w_j) = \frac{count(w_i w_j) - \delta}{count(w_i) * count(w_j)} $$
    \item
    Another interesting property of learning these distributed representations is that the word and phrase representations learned by the skip-gram model exhibit a linear structure that makes it possible to perform precise analogical reasoning using simple vector arithmetic.
  \end{itemize}


\section{Glove: Global Vectors for Word Representation} 
\label{sec:glove_global_vectors_for_word_representation}

  \textbf{Goal:}
  This paper proposes a global log-bilinear regression model that combines the advantages of the two major model families in the literature: global matrix factorization and local context window methods\cite{pennington2014glove}.\\

  \textbf{Description:}
  \begin{itemize}
    \item
    While methods like LSA efficiently leverage statistical information, they do relatively poorly on word analogy tasks, indicating a sub-optimal vector space structure. Methods like skip-gram may do better on analogy tasks, but they poorly utilize the statistics of the corpus since they train on separate local context windows instead of on global co-occurrence counts.
    \item
    The relationship between any arbitrary words can be examined by studying the ratio of their co-occurrence probabilities with various probe words.
    \item
    The authors suggest that the appropriate starting point for word vector learning should be with ratios of co-occurrence probabilities rather than the probabilities themselves.
    \item
    We can express this co-occurrence relation as shown below
    $$F((w_i - w_j)^T w_k) = \frac{P_{ik}}{P_{jk}}$$
    This makes the feature matrix interchangeable with its transpose.
    \item
    An additive shift is included in the logarithm, $$\log(X_{ik}) \Rightarrow log(1 + X_{ik})$$ which maintains the sparsity of X while avoiding the divergences while computing the co-occurrences matrix.
    \item 
    The model obtained in the paper could be compared to a global skip-gram model as opposed to a fixed window-size skip-gram model as proposed by \cite{mikolov2013efficient}.
    \item 
    The performance seems to increase monotonically with an increase in training data.
  \end{itemize}


\section{Discussion} 
\label{sec:discussion}

  Following the literature survey, this section re-visits the original research questions and provides a succinct summary that can be inferred from the experimental results and conclusions drawn from the original papers.\\

  \begin{itemize}
    \item [RQ1]
    \textbf{What are the relatively simple statistical techniques to extract features from text?} \\
    Word count frequency models like n-gram and simple bag-of-words models such as TF-IDF are still the easiest tools to obtain an numeric vector representation of text.
    \item [RQ2]
    \textbf{Is there any inherent benefit to using neural networks as opposed to the simple methods?} \\
    The benefit of using neural nets primarily is their ability to identify obscure patterns, and remain flexible enough for a varied set of application areas from topic classification to syntax parse-tree generation.
    \item [RQ3]
    \textbf{What are the trade-offs that neural networks incur as opposed to the simple methods?} \\
    The trade-offs are typically expressed in terms of computational cost and memory usage, although model complexity is a factor too, given that neural nets can be trained to learn arbitrarily complex generative models.
    \item [RQ4]
    \textbf{How do the different techniques compare to each other in terms of performance and accuracy?} \\
    This question can only be answered subjectively as it varies from application to application. Typically, document similarity can be tackled with a simple statistical approach like TF-IDF. CNNs inherently model input data in a manner that iteratively reduces the dimensionality, making it a great fit for topic classification and document summarization. RNNs are great at modeling sequences of text, which make them apt for language syntax modeling. Amongst the frameworks, GloVe's pre-trained word-embeddings perform better than vanilla Word2Vec, which is considered state-of-the-art.
    \item [RQ5]
    \textbf{In what use-cases do the trade-offs outweigh the benefits of neural networks?} \\
    As explained for the previous question, for a simple information retrieval use case such as document ranking, models such as TF-IDF, and word PMI (pointwise mutual information) are sufficient, and neural networks would be overkill in such use-cases.
  \end{itemize}


\section{Conclusion} 
\label{sec:conclusion}

  This paper has summarized the important aspects of the state-of-the-art neural network techniques that have emerged in recent years. The field of machine translation, natural language understanding and natural language generation are important areas of research when it comes to developing a range of applications from a simple chatbot, to the conceptualization of a general AI entity.

  The discussion section aggregates the results of the surveyed papers and offers a ready reference for new-comers to the field. 

  For future work, it is intended to experimentally compare different word-embedding approaches to act as a bootstrapping method to iteratively build high quality datasets for future machine learning model usage.


\section{Acknowledgments} 
\label{sec:acknowledgments}

  The author would like to thank Dr. Pascal Poupart for his constructive feedback on the survey proposal.


\bibliographystyle{acl_natbib}
\bibliography{cs698_project_report}

\end{document}